\documentclass[sigconf]{acmart}
\AtBeginDocument{%
  \providecommand\BibTeX{{%
    \normalfont B\kern-0.5em{\scshape i\kern-0.25em b}\kern-0.8em\TeX}}}

\setcopyright{acmcopyright}

\copyrightyear{2023} 
\acmYear{2023} 
\setcopyright{acmlicensed}\acmConference[SA Conference Papers
'23]{SIGGRAPH Asia 2023 Conference Papers}{December 12--15, 2023}{Sydney, NSW, Australia}
\acmBooktitle{SIGGRAPH Asia 2023 Conference Papers (SA Conference Papers '23), December 12--15, 2023, Sydney, NSW, Australia}
\acmPrice{15.00}
\acmDOI{10.1145/3610548.3618155}
\acmISBN{979-8-4007-0315-7/23/12}

\begin{document}

\title{SinMPI: Novel View Synthesis from a Single Image with Expanded Multiplane Images 
}

\author{Guo Pu}
\email{guopu@pku.edu.cn}
\affiliation{%
  \institution{Wangxuan Institute of Computer Technology, Peking University}
  \city{Beijing}
  \country{China}}

\author{Peng-Shuai Wang}
\email{wangps@hotmail.com}
\affiliation{%
  \institution{Wangxuan Institute of Computer Technology, Peking University}
  \city{Beijing}
  \country{China}}

\author{Zhouhui Lian}
\authornote{Corresponding author}
\email{lianzhouhui@pku.edu.cn}
\affiliation{%
  \institution{Wangxuan Institute of Computer Technology, Peking University}
  \city{Beijing}
  \country{China}}

\renewcommand{\shortauthors}{Pu, et al.}

\begin{abstract}

Single-image novel view synthesis is a challenging and ongoing problem that aims to generate an infinite number of consistent views from a single input image. Although significant efforts have been made to advance the quality of generated novel views, less attention has been paid to the expansion of the underlying scene representation, which is crucial to the generation of realistic novel view images. 
This paper proposes SinMPI, a novel method that uses an expanded multiplane image (MPI) as the 3D scene representation to significantly expand the perspective range of MPI and generate high-quality novel views from a large multiplane space.
The key idea of our method is to use Stable Diffusion~\cite{rombach2021highresolution} to generate out-of-view contents, project all scene contents into an expanded multiplane image according to depths predicted by monocular depth estimators, and then optimize the multiplane image under the supervision of pseudo multi-view data generated by a depth-aware warping and inpainting module.
Both qualitative and quantitative experiments have been conducted to validate the superiority of our method to the state of the art. Our code and data are available at \url{https://github.com/TrickyGo/SinMPI}.

\end{abstract}

%
%

\ccsdesc[500]{Computing methodologies~Computer graphics}
\ccsdesc[500]{Computing methodologies~Rendering}

%
\keywords{Novel view synthesis, Multiplane images, Diffusion models}

\begin{teaserfigure}
  \includegraphics[width=\textwidth]{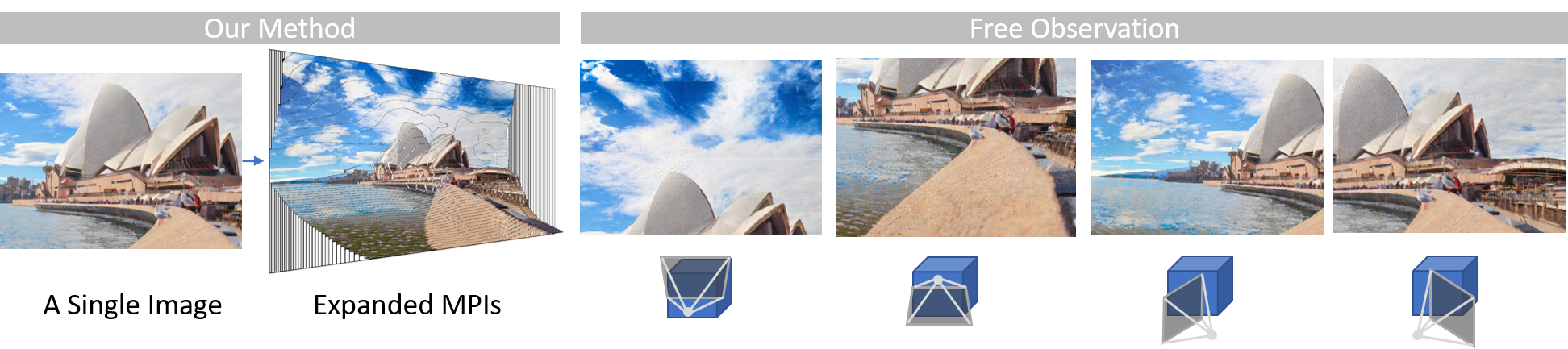}
  \caption{With only a single image as input, our method constructs its expanded multiplane image, allowing free observation from a large range of perspectives.}
  \label{fig:teaser}
\end{teaserfigure}


\maketitle

\section{Introduction}


Single-image novel view synthesis substantially narrows the gap between 2D images and 3D experiences, empowering the generation of an infinite number of views from a single input image. While the geometry of the input image can be obtained by monocular depth estimators (e.g., DPT~\cite{ranftl2021vision}), the major challenge lies in inferring the occluded content and reconstructing the missing geometry with exceedingly under-constrained 3D conditions provided by the single input image.

NeRF~\cite{mildenhall2021nerf} and its following works have demonstrated their ability to render high-quality images. However, optimizing the continuous fields of NeRF requires sufficient multi-view inputs. Due to insufficient input information, the state-of-the-art single-image NeRF-based method SinNeRF ~\cite{xu2022sinnerf} produces novel views with blurry and flickering artifacts.

With the limited information provided by a single perspective, modeling the 3D scene using discrete multi-plane representation is more efficient than continuous radiance field representations. 
MPI-based methods~\cite{shade1998layered,szeliski1999stereo,zhou2018stereo} represent the 3D scene using multiple planes in a perspective frustum, possess higher efficiency and require lower computational cost compared to methods (e.g., NeRF) using continuous 3D representations at the expense of continuous expression ability. Benefiting from the efficiency of MPIs, AdaMPI~\cite{han2022single} achieves state-of-the-art performance in single-image novel-view synthesis for images in the wild. However, MPI-based methods are constrained in their modeling space to the original camera frustum. Moreover, their rendered results may be plagued by depth discretization artifacts and repeated texture artifacts, which hinder their applications in practical use. 

To overcome these limitations, we propose SinMPI, which leverages an expanded, learnable MPI with a wider view range to represent an extended scene from a single image. Compared with previous MPI-based methods, our approach significantly enhances the capacity of modeling the 3D scene and permits a larger range of camera movements beyond the boundaries of the input image. Unlike previous MPI-based methods, which attempt to predict occluded geometry information in space using image generators which generate low-quality texture extensions, we leverage an MPI as a volumetric scene representation and fill in missing geometry supervised by pseudo-multi-view images. A key difference is that we represent our MPI as learnable parameters that are directly optimized through volume rendering, considering the expanded 3D scene is more complex and challenging to predict in a single pass. In this manner, our approach not only yields an expanded multi-plane representation that allows free observation and fast, 3D-consistent novel view synthesis, but also effectively alleviates the depth discretization artifacts and the repeated texture artifacts through volume rendering. Experimental results demonstrate that our method not only achieves the best single-image novel view synthesis results in real-world scenarios, but also possesses remarkable scene extension capabilities. 

In summary, the major contributions of this paper are as follows:

1. We introduce SinMPI, a novel single-image view synthesis method that constructs an expanded multiplane image from a single input view, enabling unrestricted observations and fast, 3D-consistent novel view synthesis. Our approach significantly extends the capacity of original MPI-based methods and supports a wider range of camera movements.

2. We leverage an MPI as a volumetric scene representation, which improves the accuracy for fitting missing geometry information and alleviates the depth discretization artifacts and the repeated texture artifacts caused by previous MPI predictors~\cite{han2022single}. \looseness=-1

3. We extensively evaluate our approach on multiple datasets, including the real-world LLFF dataset, the NeRF synthetic dataset, the DTU dataset, and the COCO dataset. Our method achieves state-of-the-art render quality on single-image novel view synthesis.

\section{Related Work}
\subsection{Single-image Novel View Synthesis}
There exist several methods that address the single-image novel view synthesis task without relying on explicit 3D representation for the entire scene, such as DISN~\cite{xu2019disn} and LearningToRecover~\cite{yin2021learning}. SynSin~\cite{wiles2020synsin} projected the input view’s point cloud to estimate new points and generate target views through a refinement network. Infinite-Nature ~\cite{liu2021infinite} and Infinite-Nature-Zero ~\cite{li2022infinitenature} generated perpetual views of natural scenes by using a differentiable mesh renderer and a GAN~\cite{goodfellow2020generative} to synthesize novel views. 4D-RGBD-Light Field~\cite{srinivasan2017learning} synthesized a light field from an image. Pixelsynth~\cite{rockwell2021pixelsynth} used point cloud as an intermediate representation and then predicted novel views with CNN decoders. 3DiM~\cite{watson2022novel} proposed a geometry-free pose-conditional image-to-image diffusion model.

However, these methods did not employ explicit 3D representations for entire scenes, which limits the 3D view consistency of the generated views and produces flickers and blurs in the synthesized results, especially in occluded regions. In contrast, our method ensures multi-view consistency using a fixed MPI for the entire scene.  \looseness=-1

\subsection{NeRF-based Methods}
NeRF~\cite{mildenhall2021nerf} and its variants have received significant attention in novel view synthesis. However, the original NeRF required at least 100 posed views for input, which was both inconvenient and inaccessible for practical applications. To overcome this challenge, several NeRF-based methods were proposed to generate novel views based on fewer input images, including PixelNerf~\cite{yu2021pixelnerf}, DietNeRF~\cite{jain2021putting}, Pix2nerf~\cite{cai2022pix2nerf}, PortraitNeRF~\cite{gao2020portrait}, ViTNeRF~\cite{lin2023vision}, NerfDiff~\cite{gu2023nerfdiff} and Nerdi~\cite{deng2023nerdi}. For scene-level modeling, SinNeRF~\cite{xu2022sinnerf} is the state-of-the-art NeRF-based method for single-image novel view synthesis. 

The advantages of NeRF lie in its ability to simultaneously model diffuse and specular color in multi-view images. However, inferring a light field  from a single image input is challenging for NeRF. The difficulty of optimizing NeRF from a single input view hampers the ability of NeRF-based methods to generate photo-realistic results. In contrast, MPI, as a diffuse volumetric scene representation, is more efficient and sufficient for tasks where light field information is not required. Nevertheless, the potential of NeRF in modeling view-dependent effects for this task is also worth exploring.

\subsection{Layer-based Methods}
Layer-based methods represent a 3D scene using discrete layers, resulting in higher efficiency and lower computational cost compared to continuous 3D representations like NeRF. Layer-based methods, such as 3D-photography~\cite{shih20203d} and SLIDE~\cite{jampani2021slide}, can produce high-quality synthesis results from a single input image. 

As a discrete diffuse volumetric scene representation, MPI-based methods~\cite{shade1998layered,szeliski1999stereo,zhou2018stereo} were proposed to represent the scene using multiple planes in a perspective frustum. Following multi-view MPI-based works~\cite{srinivasan2019pushing,wizadwongsa2021nex} improved the MPIs render quality. Srinivasan et al.~\shortcite{srinivasan2019pushing} proposed a two-step MPI prediction procedure to alleviate issues of depth discretization artifacts and repeated texture artifacts that appear in MPIs rendering results. NeX~\cite{wizadwongsa2021nex} extended MPIs for non-Lambertian material and was able to handle view-dependent effects effectively. DeepView~\cite{flynn2019deepview} incorporated occlusion reasoning to generate an MPI from a set of sparse camera viewpoints. For single-image novel view synthesis, following MPI-based methods (SVMPI~\cite{tucker2020single}, VMPI~\cite{li2020synthesizing}, MINE~\cite{li2021mine}, GMPI~\cite{zhao2022generative}, and AdaMPI~\cite{han2022single}) trained convolutional image generators to predict missing occluded geometry in the space. However, these methods heavily rely on large-scale datasets to train MPI predictors; thus, the MPI prediction quality is highly constrained by the training data and can hardly adapt to new manifolds, such as synthetic object-centric scenes in Fig.~\ref{fig: fig_exp_5}.

The typical MPIs renderer only models a frustum range, resulting in a limited view field that hinders the practical application. To address this issue, we propose a novel method that uses larger planes in space to expand the view range. Additionally, MPI-based methods’ rendering results often suffer from depth discretization artifacts and repeated texture artifacts. We alleviate these issues by optimizing an MPI through volume rendering.

\section{Method}
\begin{figure*}[h]
  \includegraphics[width=1.0\textwidth]{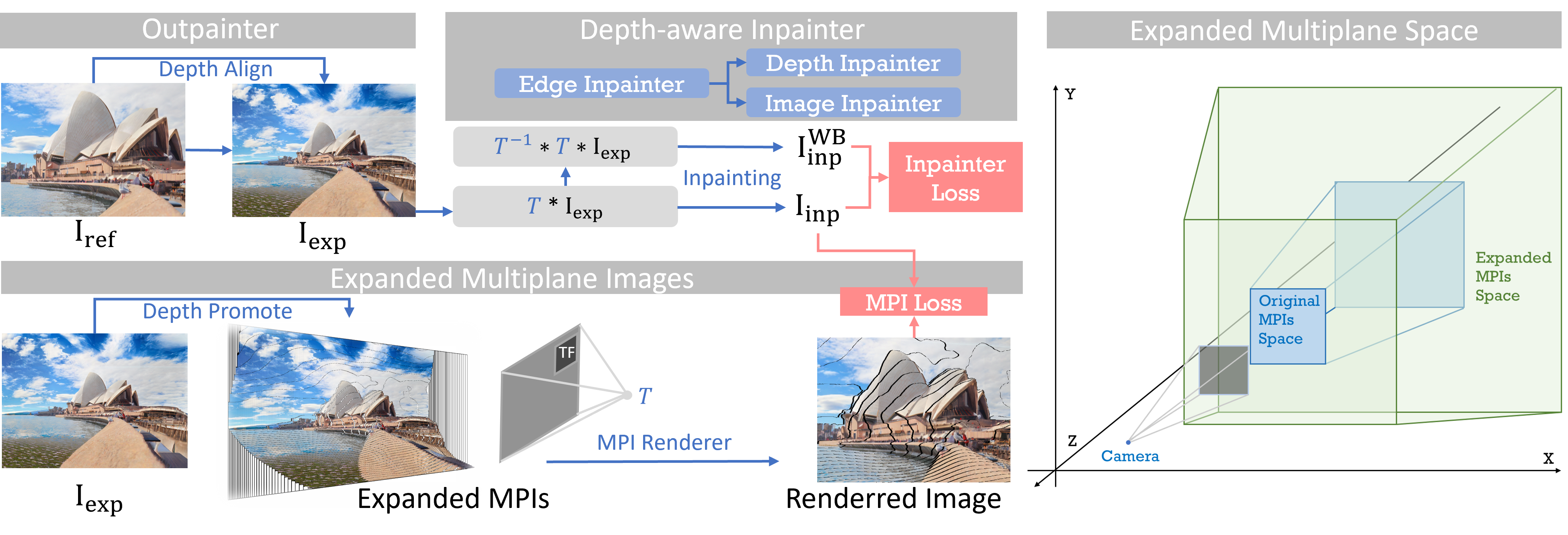}
  \caption{The pipeline of our SinMPI consists of three major components: a continuous outpainter, a depth-aware inpainter, and an expanded MPI. We first extrapolate out-of-scene contents, infer the scene's 3D geometry and occluded contents, and finally convert the scene into a multiplane space. Note that trainable filters (TF) are used to aggregate local rays. On the right, we show our expanded multiplane space. To enable the construction of larger 3D spaces, we employ larger planes within a larger frustum (green) than the camera frustum (blue) in our expanded MPI space. This approach allows for larger camera view ranges, including larger rotation angles than those supported by existing MPI-based methods.}
  \label{fig:fig2}
\end{figure*}
In this section, we present the details of our SinMPI, which expands the perspective range of an MPI and generates high-quality 3D-consistent novel views. 
The key idea is to generate out-of-view contents via Stable Diffusion~\cite{rombach2021highresolution}, project them into a multiplane image, then optimize the MPI under the supervision of pseudo-multi-view images generated by a depth-aware warping and inpainting module. Fig.~\ref{fig:fig2} illustrates the pipeline of our method, and additional details are presented in the following subsections.  \looseness=-1

\subsection{Continuous Outpainting Module}
We utilize Stable Diffusion~\cite{rombach2021highresolution} to perform continuous extrapolation of user-provided reference images for scene expansion. The inpainting model based on Stable Diffusion, leverages semantic masks as a conditioning input, and the conditional latent diffusion model is trained by minimizing:
\begin{equation}
    L_{LDM} := \mathbb{E}_{\varepsilon(x),y,\epsilon\sim \mathcal{N}(0,1),t}[\left\| \epsilon-\epsilon_{\theta}(z_{t},t,m) \right\|^{2}_{2}],
\end{equation}
where $t = 1, 2, ... T$ denotes the time step, $z_{t}$ is the noisy version of the latent vector $z$ of input $x$, $m$ represents the mask, $\epsilon$ is the noise schedule, and $\epsilon_{\theta}$ denotes the time-conditioned UNet, respectively.

For unposed single images, to maintain consistency with the input view’s depth range, we estimate the depth of the reference image using a state-of-the-art depth estimator, DPT~\cite{ranftl2021vision}, both before and after the outpainting process. Our depth alignment is a linear scale and shift of the outpainted monocular depth which preserves the depth range of the original areas. We align the depth of the original region accordingly, resulting in the reference image $I_{ref}$, its depth $D_{ref}$, the expanded image $I_{exp}$, and its depth $D_{exp}$. Importantly, we ensure that the geometry in the original view remains in the same depth range and the pre-defined camera trajectories are unchanged. 

\subsection{Depth-aware Inpainting Module}
To reconstruct a complete 3D scene from the outpainted image and its aligned depth, we need to infer the occluded content. We start by treating the pixels of the reference view, along with their depth values, as a point cloud in space. We then use depth-warp~\cite{huang2021m3vsnet} to warp the original image to novel viewpoints, which involves computing:
\begin{equation}
p_{j}=\hat{K}T(K^{-1}D_{i}p_{i}),
\end{equation}
where $p$ denotes the pixel point, $i$, and $j$ denote the indices of the source and target views, respectively, $T$ is the camera extrinsic, $D$ represents the depth, and $K$ and $\hat{K}$ denote the source and target camera intrinsics, respectively.

To generate projected views, we render the warped point cloud with the Painter’s Algorithm
~\cite{newell1972solution} to choose points nearest to the camera. This process creates novel views with missing areas that require proper inpainting to obtain a complete novel view.

For inpainting, we draw inspiration from 3D-photography~\cite{shih20203d} and AdaMPI~\cite{han2022single} and use a depth-aware inpainter based on Edge-connect~\cite{nazeri2019edgeconnect} which is constructed by an edge inpainter, followed by a depth inpainter and an image inpainter. The inpainter is pretrained on the Places2 dataset~\cite{zhou2017places} and then we deploy a per-scene fine-tuning on the input image $I_{exp}$.

To fine-tune the inpainter on a single image, we first warp the image $I_{exp}$ to a novel view $T * I_{exp}$ according to a sampled camera $T$. We then inpaint the warped image to obtain the inpainted image $I_{inp}$. Since there is no ground truth available for $I_{inp}$, we utilize deep features extracted from specific layers of a pre-trained VGG network for texture matching at the feature/patch level. The perceptual loss is defined as:
\begin{equation}
L_{per}(I_{inp}, I_{exp}) = \frac{1}{N_l}\sum_{x=1}^{W_l}\sum_{y=1}^{H_l}\sum_{z=1}^{C_l}\parallel \phi_l(I_{inp})_{x,y,z}-\phi_l(I_{exp})_{x,y,z}  \parallel _1,
\end{equation} 
where $\phi_l$ denotes the output feature of the layer $l$ of the VGG19 network, $N_l$ equals $W_{l}H_{l}C_{l}$,  and $W_{l}, H_{l}$ and $C_{l}$ represent the spatial width, height, and depth of the feature $\phi_l$, respectively.

In addition to the VGG-based perceptual loss, we also incorporate the features from a vision transformer to calculate the ViT loss:
\begin{equation}
L_{vit}(I_{inp}, I_{exp}) = \parallel f_{vit}(I_{inp}) -  f_{vit}(I_{exp})\parallel^{2},
\end{equation}
where $f_{vit}$ denotes the feature extracted by  DINO-ViT~\cite{caron2021emerging} pre-trained on the ImageNet~\cite{deng2009imagenet} dataset.

Next, we warp the image $T * I_{exp}$ back by computing $T^{-1} * T * I_{exp}$ and then obtain the inpainted warp-back image $I^{WB}_{inp}$. The warp-back depth is also inpainted to get $D^{WB}_{inp}$.  To ensure that the reconstructed image and depth are accurate, we apply the reconstruction loss, along with the VGG/ViT feature/patch level losses, to the reconstructed image. The reconstruction loss is defined as:
\begin{equation}
L_{rec}(I^{WB}_{inp}, I_{exp}) = \parallel I^{WB}_{inp} - I_{exp}\parallel_{1}.
\end{equation}
To enforce depth consistency, we apply the depth loss to the inpainted depth:
\begin{equation}
L_{depth}(D^{WB}_{inp}, D_{exp}) = \parallel D^{WB}_{inp} - D_{exp}\parallel_{1},
\end{equation}
where $D_{exp}$ and $D^{WB}_{inp}$ represent the depth map of the expanded image and the inpainted warp-back depth map, respectively.

The total loss for finetuning the inpainter is:
\begin{equation}
\begin{aligned}
&L_{inpainter}(I_{inp}, D_{inp}, I^{WB}_{inp}, D^{WB}_{inp}, I_{exp}, D_{exp}) = \\
&\lambda_{rec}L_{rec}(I^{WB}_{inp}, I_{exp}) +
\lambda_{depth}L_{depth}(D^{WB}_{inp}, D_{exp}) + \\
&\lambda_{per}L_{per}(I^{WB}_{inp}, I_{exp}) + \lambda_{vit}L_{vit}(I^{WB}_{inp}, I_{exp}) + \\
&\lambda_{per}L_{per}(I_{inp}, I_{exp}) + \lambda_{vit}L_{vit}(I_{inp}, I_{exp}),
\end{aligned}
\end{equation}
where $\lambda_{rec}$, $\lambda_{depth}$, $\lambda_{per}$, and $\lambda_{vit}$ denote the loss weights. We experimentally found that finetuning the inpainting model for around 100 iterations per view strikes a good balance between underfitting and overfitting.

Using our finetuned depth-aware warping and inpainting module, we generate pseudo-multi-view data by first warping the source point to target views via depth warping and then inpainting the holes of warped images from different perspectives of the 3D scene. Note that the generated pseudo-novel views may contain certain pixels that are not fully 3D-consistent due to the lack of explicit 3D geometry information in the inpainted areas. To address this issue, we use pseudo-novel views to optimize a single MPI, thereby enabling the creation of a complete and static 3D representation. \looseness=-1

\subsection{Expanded MPI}
As information from a single view is inadequate to fully capture a 3D scene, we employ discrete multiplane representations, which are efficient and feasible for the task of synthesizing novel views from a single image. Due to the substantially reduced computational costs compared to continuous representations, our method can efficiently construct large 3D spaces and generate photo-realistic novel views that span a large view range. To accomplish this, we first project all contents onto our expanded multiplane image and then optimize the multiplane image using previously-generated pseudo-multi-view images as supervision.

\subsubsection{MPI Initialization}

Since the points of the original view provide the ground-truth information in the 3D space for single-image novel view synthesis, they should remain fixed. Therefore, we segment the original image according to depth and project it onto multiple planes evenly distributed along the depth. Each plane stores only a small range of the depth space and has its initialized contents frozen in the following procedures. Our expanded MPI contains $P * H * W * 4$ parameters, where $P$ denotes the number of planes, $H * W$ represents the size of a plane, and $4$ is the channel number of a point in a plane, which stores the point’s RGB values and density. \looseness=-1

\subsubsection{MPI Renderer}

The original method using MPIs represents the 3D scene as a series of parallel planes in a view frustum. The plane homography transformation~\cite{hartley2003multiple} is defined by:
\begin{equation}
    \left[ x_{i},y_{i},1 \right]^{T} \sim K(R-\frac{tn^{T}}{d_{k}})K^{-1} \left[ x_{j},y_{j},1 \right]^{T},
\end{equation}
where $n = [0, 0, 1]^{T}$ is the plane normal, $d_k$ is the depth of the plane $k$, $x$ and $y$ denote the coordinates of a pixel in the target view $j$ or source view $i$, respectively, $K$ denotes the camera intrinsics, and $R$ and $t$ are the rotation and translation matrices, respectively.

The layers of our MPI are distributed evenly along the $z$-axis of the entire 3D space. We render the target view through volume rendering using the following equations: 
\begin{equation}
I = \sum_{i=1}^{N}T_{i}(1- {exp}(-\sigma_{d_{i}}\delta_{d_{i}}))c_{d_{i}},
\end{equation}
\begin{equation}
T_{i} = {exp}(-\sum_{j=1}^{i-1}(-\sigma_{d_{i}}\delta_{d_{i}})),
\end{equation}
where $T$ denotes the accumulated transmittance starting from the first plane, $c$ denotes the RGB value, and $\sigma$ represents the density. Each plane $i$ is considered as a bin along the depth $d$, and $\delta$ is the distance between every two planes, which is a constant in our case due to the even distribution of planes along the depth.

The original method using MPIs simply represents the 3D scene in a view frustum. Once the camera rotates or translates to a certain amount, it can easily exceed the field of the original MPIs' view range. Current MPI-based methods~\cite{li2021mine, han2022single} typically use border pixel extension for out-of-scene contents which generates unnatural textures. This inherent limitation makes the MPIs representation less applicable to real-world scenarios with significant camera movement.

In order to endow MPIs with a wider modeling space, we propose to use larger planes for the view frustum with an increased view angle of $\theta$, while maintaining the camera’s original field of view (see the right side of Fig.~\ref{fig:fig2}). The relationship between the increased  frustum angle and the expanded plane size is defined by: $\frac{w * a}{2f}=tan(\frac{\theta}{2})$, where $w$ represents the length of the long plane border, $a$ denotes the expansion factor, $f$ is the focal length of the camera, and $\theta$ is the view angle range, respectively. During the testing phase, we perform homography transformation on the expanded MPI and only apply volume rendering to the parameters within the camera’s field of view instead of all the planes.

\subsubsection{Local Rays Aggregation}
Unlike existing MPI-based methods that predict MPI through convolutional image decoders, we directly optimize the MPI itself. This novel approach allows for improved optimization efficiency and faster convergence rates since the optimization of each ray is independent. Due to the dense nature of the point cloud representation and the limitation of the inpainting module, our created pseudo views are sharp and prone to noise points as well as flickering artifacts. Alleviating these artifacts requires patch-level supervision. 

To facilitate interaction among local rays and create smooth rendered images, we introduce a convolutional filter layer after the MPI renderer to enable sharing of information within local patches and facilitate faster optimization. Specifically, we use a trainable bilateral filter~\cite{tomasi1998bilateral} to aggregate local rays by weighing the influence of neighboring rays based on their intensities and locations. For an MPI rendered image $I_{mpi}$, we perform the following discrete filter operation:

\begin{equation}
\hat{I}_{mpi}(p) = \frac{1}{W_{p}}\sum_{q\in S}^{}G_{\delta_{s}}(\parallel p-q\parallel)G_{\delta_{r}} (|I_{mpi}(p)-I_{mpi}(q)|)I_{mpi}(q),
\end{equation}
with the normalization factor:
\begin{equation}
W_{p} = \sum_{q\in S}^{}G_{\delta_{s}}(\parallel p-q\parallel)G_{\delta_{r}} (|I_{mpi}(p)-I_{mpi}(q)|),
\end{equation}
where $\hat{I}_{mpi}$ is the filtered image, $p$ is the pixel index, $q$ is the index in patch $S$, $G_{\delta_{s}}$ is the learnable Gaussian kernel, $G_{\delta_{r}}$ is the pixel difference kernel, respectively.

To weigh the intensity difference between the center pixel and other pixels, the pixel difference kernel $G_{\delta_{r}}$ is calculated as:
\begin{equation}
G_{\delta_{r}} (|I_{mpi}(p)-I_{mpi}(q)|) = e^{-\frac{[I_{mpi}(i,j) - I_{mpi}(m,n)]^{2}}{2\sigma^{2}}},
\end{equation}
where $i$ and $j$ are the center indices in the patch, $m$ and $n$ are the pixel coordinates in the center index of a patch, and $\sigma$ is the kernel width, respectively. 

To weigh the location difference between the center pixel and other pixels, the learnable Gaussian kernel $G_{\delta_{s}}$ is initialized as:

\begin{equation}
G_{\delta_{s}}(\parallel p-q\parallel) = e^{-\frac{(i-m)^2 + (j-n)^2}{2\sigma^{2}}}.
\end{equation}

To make the filtered image smoother without sacrificing quality and reduce the blurry effects of the Gaussian kernel, we make $G_{\delta_{s}}$ trainable as a convolution weight, thereby making the aggregation layer adaptive.

\subsubsection{MPI Optimization}
As our expanded MPI is specifically designed for single-scene novel view synthesis, we choose to construct them as learnable parameters instead of using an MPI predictor, such as those in \cite{li2021mine, han2022single}. We found that training large-scale predictors to predict our expanded MPI is both unnecessary and computationally unreasonable.

We optimize the MPI supervised by pseudo-novel-view data created using our depth-aware warping and inpainting module. The full loss function consists of the pixel loss and the feature loss:
\begin{equation}
\begin{aligned}
&L_{MPI}(\hat{I}_{mpi},I_{inp}) = \lambda_{rec}L_{rec}(\hat{I}_{mpi},I_{inp}) + \\ &\lambda_{per}L_{per}(\hat{I}_{mpi},I_{inp}) + \lambda_{vit}L_{vit}(\hat{I}_{mpi},I_{inp}),
\end{aligned}
\end{equation}
where the $\lambda$s are the loss weights.

\section{Experiments}
In this section, we present results that demonstrate the expanded view range and diverse applications of our method. We then conduct a comprehensive quality comparison with state-of-the-art methods to evaluate the performance of SinMPI. In addition, we provide deeper insights into our approach through an ablation study that dissects the contributions of different components in our pipeline.

For more details of our continuous outpainting procedure, implementation specifics, DTU dataset evaluations and efficiency comparisons, please refer to the supplemental material.

\subsection{Applications}
Our method performs high-quality 3D-consistent novel-view synthesis from a single image. Fig.~\ref{fig: fig_exp_4} shows novel views of various expanded scenes generated by our method.

\begin{figure}[ht]
  \includegraphics[width=0.5\textwidth]{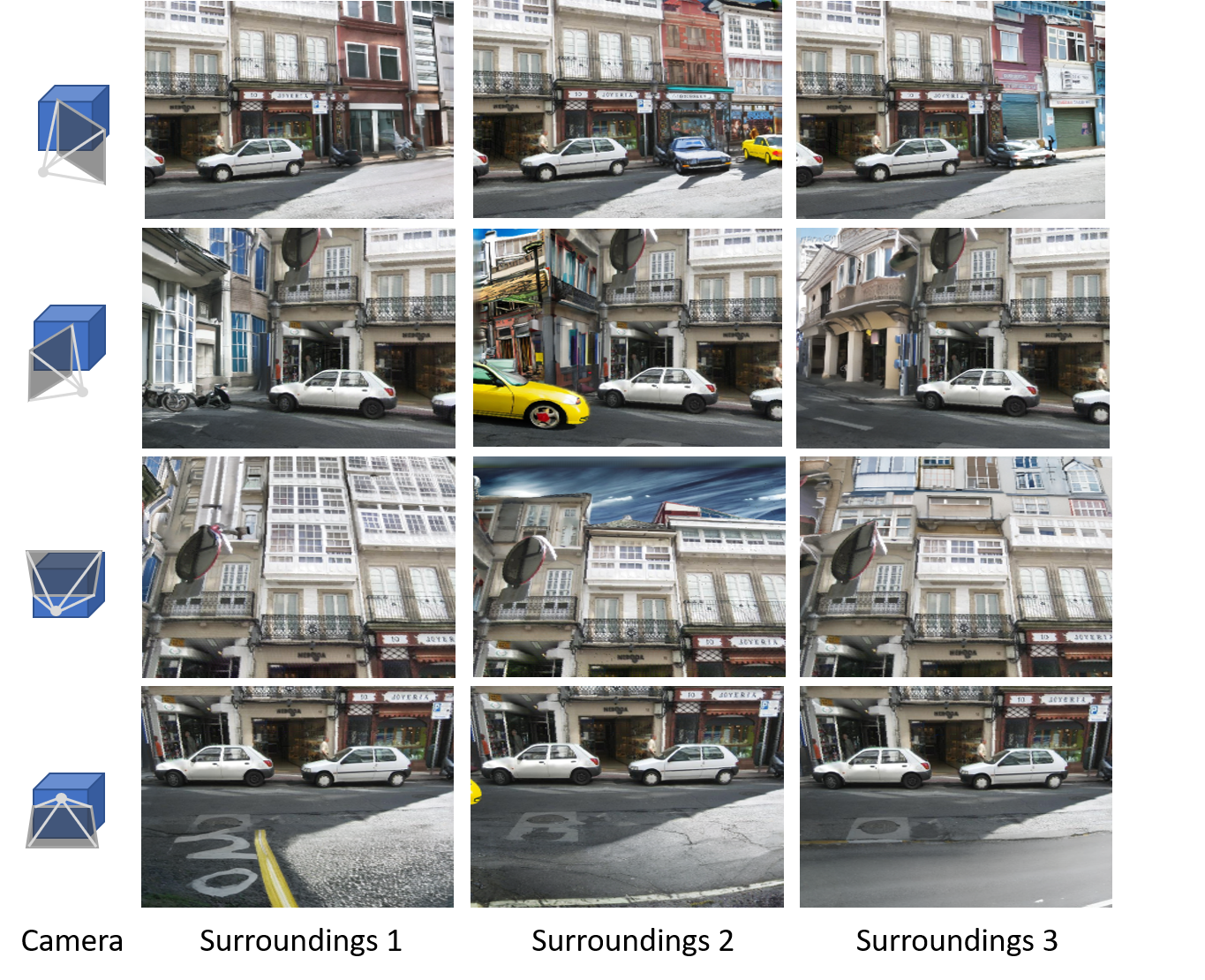}
  \caption{Single-image novel view synthesis results obtained by our SinMPI for a street scene with three different outpainted surroundings.}
  \label{fig: fig_exp_3}
\end{figure}

Furthermore, our method can integrate seamlessly with image editing approaches, enabling users to edit 3D scenes. Fig.~\ref{fig: fig_exp_3} showcases the capabilities of SinMPI, which can synthesize novel views of a street scene and alter its surroundings based on Stable Diffusion’s~\cite{rombach2021highresolution} various outpainting outputs.

For all synthesis results presented in the figures, we recommend zooming in for a more detailed inspection. We also highly recommend viewing the supplemental video for a comprehensive 3D-aware inspection that provides a better understanding of the spatial relationships and geometry of the synthesized scenes.

\subsection{Expanded View Range}
Our scene extrapolating pipeline significantly expands the scene capacity of the original method that adopts MPIs. We show the differences between ours with the expanded scene and the state-of-the-art MPI-based method AdaMPI~\cite{han2022single} in Fig.~\ref{fig: fig_exp_2}. Our method generates more photo-realistic results, especially on unseen surrounding regions. Our MPI expansion strategy can be readily applied to any MPI-based method and pushes the bounds of the MPI-based approach’s scene capacity. Nevertheless, due to the high computational cost, our scene-expanding pipeline currently cannot be integrated with other existing approaches to synthesize novel views with the same resolution as our method.

\subsection{Comparison with Previous Methods}

\subsubsection{Baselines}
We conduct a comprehensive comparison of SinMPI to several state-of-the-art methods for single-image novel view synthesis, including SinNeRF~\cite{xu2022sinnerf} DietNeRF~\cite{jain2021putting}, PixelNeRF~\cite{yu2021pixelnerf} and DS-NeRF~\cite{deng2022depth}, and the state-of-the-art MPI-based single-image novel view synthesis method AdaMPI~\cite{han2022single}. AdaMPI uses a pre-trained model on large datasets (e.g., COCO~\cite{caesar2018coco}) for scene prediction, while SinMPI is a per-scene modeling method akin to 3D reconstruction approaches like NeRF, where we train on individual images (scenes) for evaluating novel view renderings.

Comparing these baselines, we observe that AdaMPI produces higher-quality in-fov view synthesis results but SinNeRF's synthesis results possess a larger view range. However, SinMPI can synthesize novel views with both higher quality and a larger view range.

In our qualitative demonstrations, we mainly compare our method with SinNeRF~\cite{xu2022sinnerf} and AdaMPI~\cite{han2022single}, since these methods have demonstrated their superiority over other methods in the task of single-image novel view synthesis. Therefore, we omit the demonstrations of other methods' results and only report their quantitative scores for completeness.

\subsubsection{Datasets}
We compare rendering quality on the Local Light Field Fusion(LLFF) dataset~\cite{mildenhall2019local}, the NeRF synthetic dataset~\cite{mildenhall2021nerf}, and the DTU dataset~\cite{jensen2014large}. 
These three datasets contain multi-view data that can be used for evaluating both image quality and 3-D consistency. We also compare the performance of different methods by using images contained in COCO~\cite{caesar2018coco} and ADE20K~\cite{zhou2017scene}. Note that these two datasets simply consist of individual images without ground-truth multi-view data so only qualitative comparisons are available.

\subsubsection{Evaluation Protocol}
To conduct a fair comparison, for all methods including ours, the resolutions of input images are identical and their depths are estimated using the same pre-trained multi-view model. Following SinNeRF~\cite{xu2022sinnerf}, the depth maps used for the NeRF synthetic dataset and LLFF dataset are predicted by a pre-trained NeRF and the depth maps used for the DTU dataset are predicted by MVSNet. Notice that all the methods can generate novel views with the monocular predicted depth such as the DPT~\cite{ranftl2021vision} predicted depth, but in order to faithfully evaluate the generated views the depth should match the scene scale so that the generated views are aligned with the corresponding ground-truth images. 

For all quantitative comparisons with other methods, we do not use Stable Diffusion to outpaint the input image to keep inputs for all methods the same. We would like to emphasize that although our method generates more photorealistic extrapolated scenes when using outpainted images as input, the matching scores with ground-truth novel views are reduced compared to those using original images, indicating a greater mismatch with the ground-truth multi-view images introduced by the outpainting process, including scene scale mismatch due to monocular depth re-estimation and randomness in the outpainted areas.

\subsubsection{Evaluation Metrics}
We assess performance using some commonly used image quality evaluation metrics, including SSIM, PSNR, and LPIPS~\cite{zhang2018unreasonable}.

\subsection{Evaluation on the LLFF Dataset}
\subsubsection{Qualitative Comparison}
The LLFF dataset consists of real-world forward-facing scenes. For each scene, we randomly choose a posed view as the reference view and the remaining views are used as the test set.

\begin{figure}[ht]
  \includegraphics[width=0.5\textwidth]{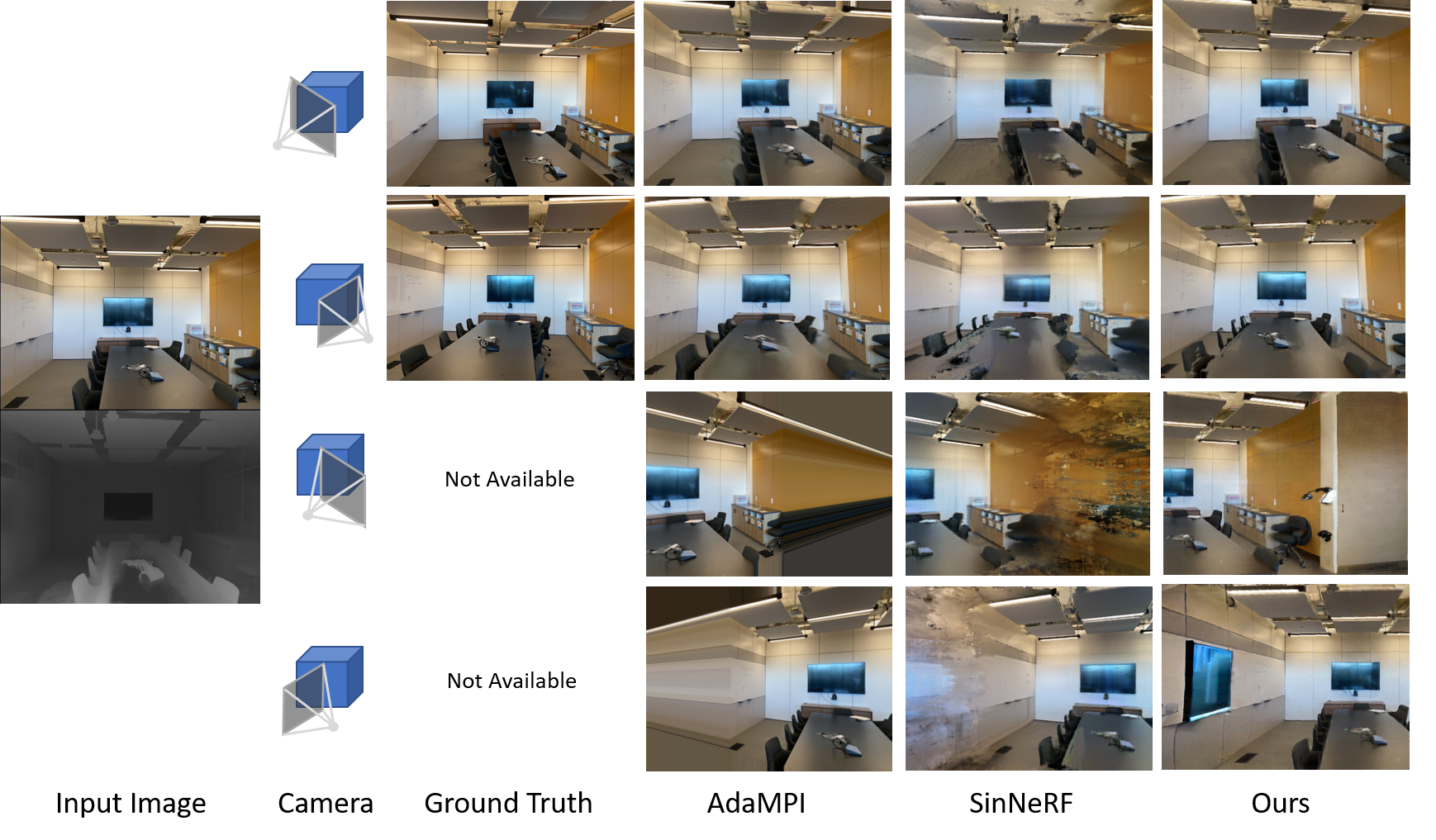}
  \caption{Qualitative results of our method against other state-of-the-art methods evaluated on the LLFF dataset (Room scene).}
  \label{fig: fig_exp_1}
\end{figure}

As shown in the top two rows of Fig.~\ref{fig: fig_exp_1}, our method produces more photorealistic results with higher resolution compared to SinNeRF, due to the layer-based representation’s greater computational efficiency and ability to faithfully preserve the original image’s resolution and textures. Our per-scene optimization and volume rendering strategy empowers our method to surpass the performance of AdaMPI.
In the last two rows of Fig.~\ref{fig: fig_exp_1}, we illustrate that SinMPI performs significantly better than previous methods when equipped with our scene extrapolating strategy, providing a much broader view range.

\subsubsection{Quantitative Comparison}
Table~\ref{tab: LLFF} compares the quantitative results of three different methods evaluated on the LLFF dataset. We observe that our method performs better than AdaMPI on the Flower and Room subsets and outperforms SinNeRF on the Room subset. 
It should be pointed out that the Flower subset’s multi-view data contains up to 19$\%$ out-of-view pixels, but MPI-based methods (i.e., AdaMPI and Ours w/o outpainting) cannot generate out-of-view pixels, which negatively impacts the evaluation scores. Nevertheless, the high-quality synthesis results produced by our method are easily noticeable in the supplemental video. 
While it is evident from our rendered results that our method produces sharper outputs which match the ground truth better compared to SinNeRF, SinNeRF has the best PSNR score, mainly due to the fact that blurry or repetitive patterns of SinNeRF react positively for the PSNR score.
Our method achieves better scores against AdaMPI, which indicates the best generated texture quality of our method.

\begin{table}[]
\caption{Quantitative comparison of our method against other state-of-the-art methods evaluated on the LLFF dataset.}
\centering
\scalebox{0.9}{
\begin{tabular}{lllllll}
\hline
          & \multicolumn{3}{l}{Room}                            & \multicolumn{3}{l}{Flower}                          \\ \cline{2-7} 
Method    & SSIM$\uparrow$ & PSNR$\uparrow$ & LPIPS$\downarrow$ & SSIM$\uparrow$ & PSNR$\uparrow$ & LPIPS$\downarrow$ \\ \hline
DS-NeRF   & 0.65           & 17.44          & 0.40              & 0.41           & 16.92          & 0.39              \\
DietNeRF  & 0.49           & 15.77          & 0.75              & 0.20           & 13.35          & 0.75              \\
PixelNeRF & 0.41           & 12.88          & 0.76              & 0.19           & 13.20          & 0.63              \\
SinNeRF   & 0.67           & \textbf{18.85} & 0.38              & \textbf{0.41}  & \textbf{17.20} & \textbf{0.37}     \\
AdaMPI    & 0.62           & 13.45          & 0.51              & 0.32           & 12.70          & 0.53              \\
Ours      & \textbf{0.70}  & 17.20          & \textbf{0.34}     & 0.38           & 15.57          & 0.38              \\ \hline
\end{tabular}
}
\label{tab: LLFF}
\end{table}

\subsection{Evaluation on the NeRF Synthetic Dataset}
\subsubsection{Qualitative Comparison}
The NeRF synthetic dataset contains images focusing on delicate objects with large camera rotations. For each scene of this dataset, we randomly choose a posed image as the reference view. To align with SinNeRF~\cite{xu2022sinnerf}, the test set consists of 60 posed images that are rendered in Blender by rotating the reference view's camera pose around y-axis in a degree range of $[-30,30]$.

\begin{figure}[ht]
  \includegraphics[width=0.5\textwidth]{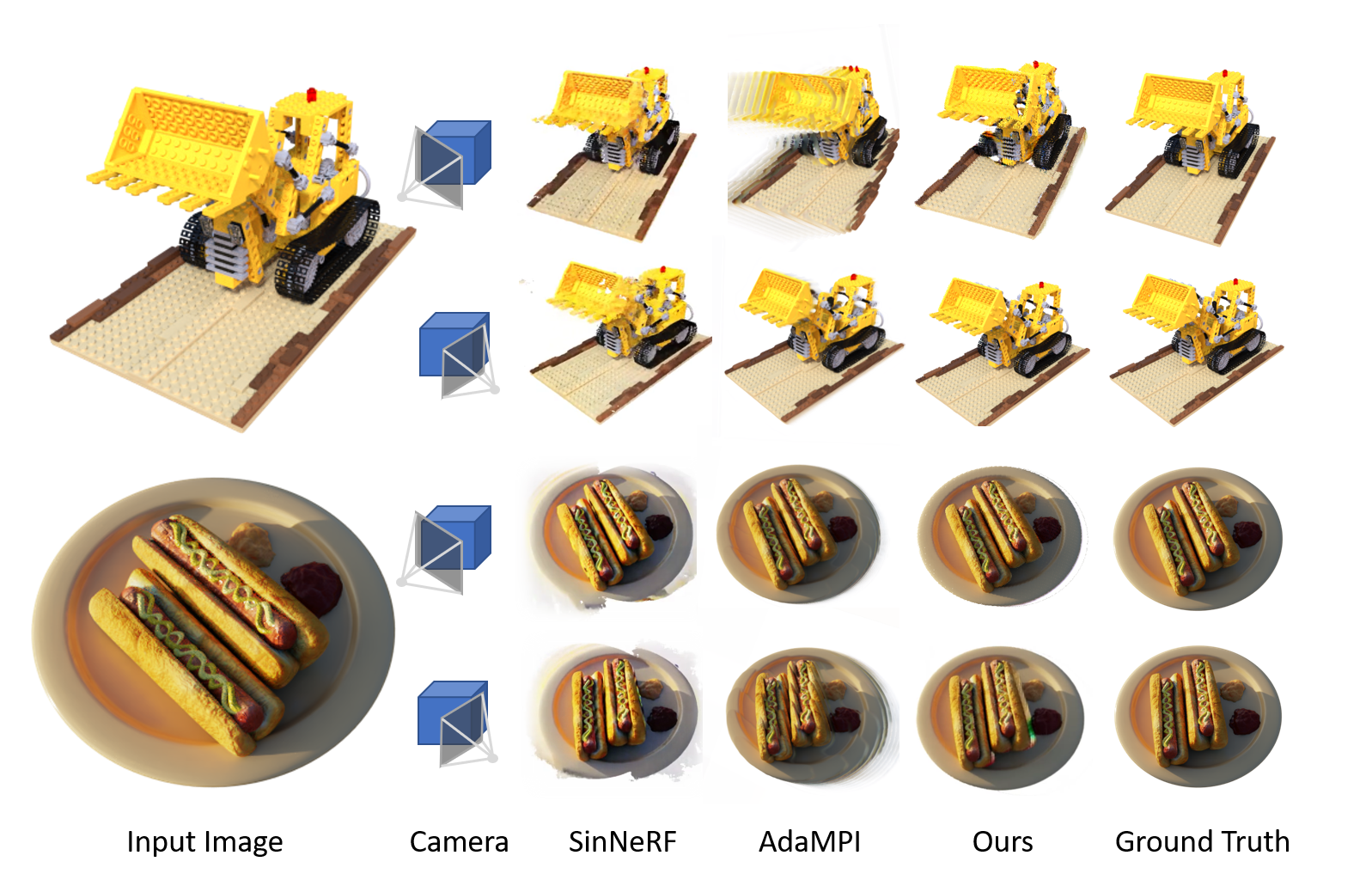}
  \caption{Qualitative results of our method against other state-of-the-art methods evaluated on the NeRF synthetic dataset (Lego scene and Hotdog scene).}
  \label{fig: fig_exp_5}
\end{figure}

\begin{figure}[ht]
  \includegraphics[width=0.5\textwidth]{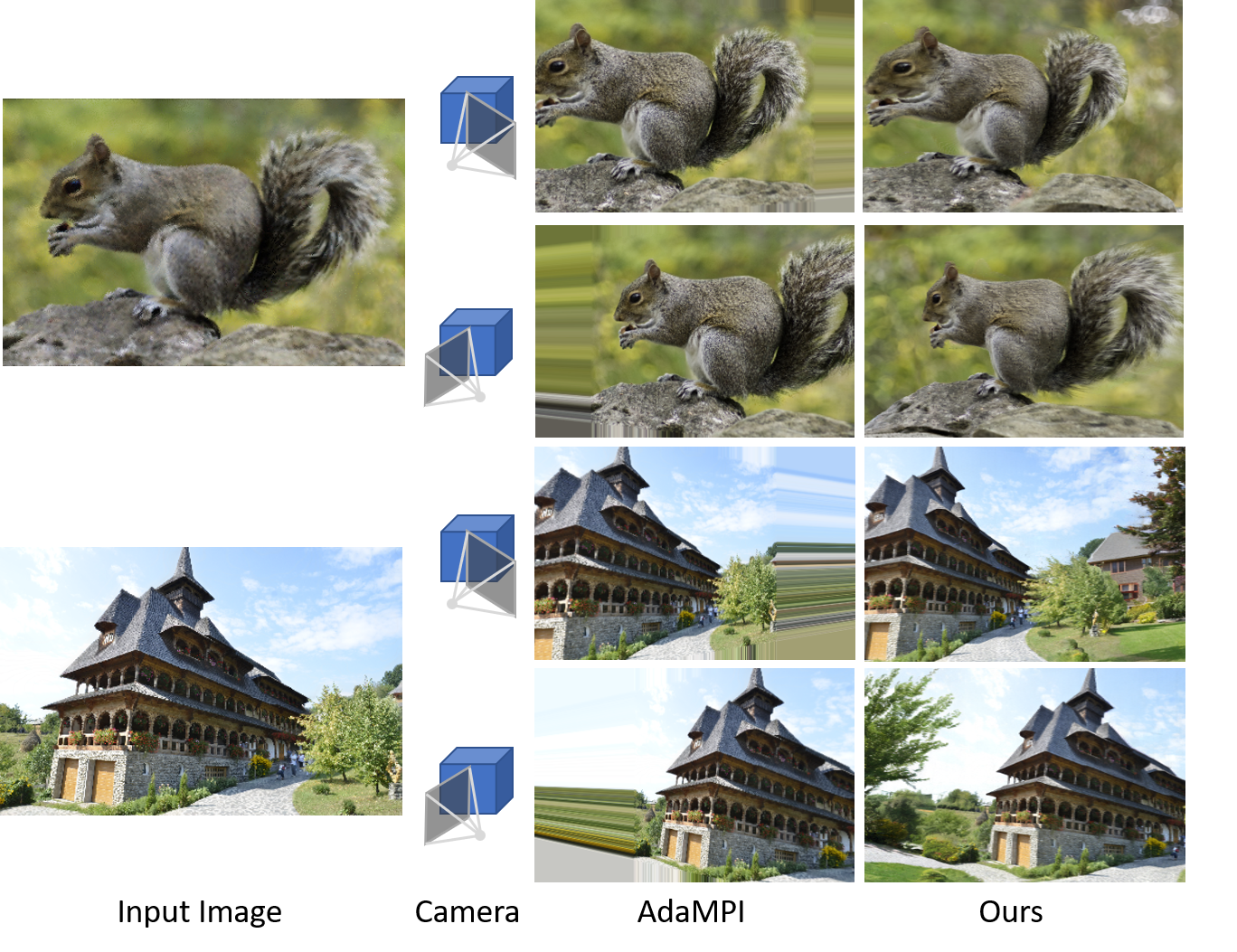}
  \caption{The demonstration of our expanded MPI supporting large camera rotation with the expanded scene. Our proposed MPI expanding strategy pushes the boundary of MPI-based methods with a larger scene capacity.}
  \label{fig: fig_exp_2}
\end{figure}

\begin{figure*}
  \includegraphics[width=0.9\textwidth]{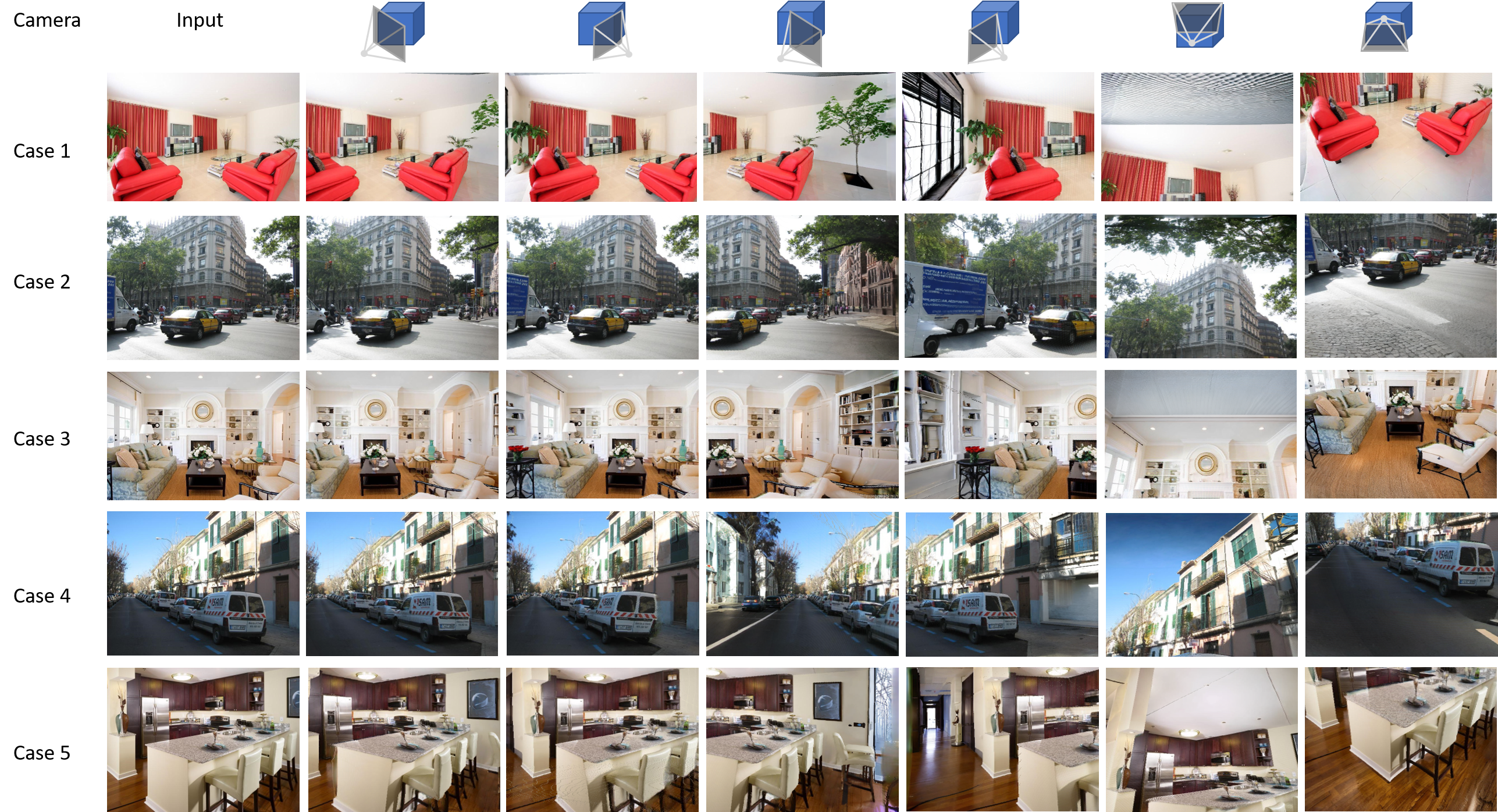}
  \caption{Single-image novel view synthesis results obtained by our SinMPI for various scenes.}
  \label{fig: fig_exp_4}
\end{figure*}

Fig.~\ref{fig: fig_exp_5} shows the qualitative comparison of SinNeRF and our SinMPI evaluated on the NeRF synthetic dataset. SinNeRF’s results exhibit more noise due to MLP underfitting from limited one-view information. On the other hand, AdaMPI’s results are prone to heavy depth discretization artifacts and repeated texture artifacts caused by its MPI predictors.
Our method, in contrast, generates novel-view images with high-quality geometry. We achieve this by leveraging MPIs as a volumetric scene representation, which enhances the accuracy of fitting missing geometry information and mitigates depth discretization artifacts and repeated texture artifacts.  \looseness=-1

\subsubsection{Quantitative Comparison}
We also quantitatively compare the performance of SinNeRF and our method on the NeRF synthetic dataset. Table~\ref{tab: synthetic} presents the comparison results, which demonstrate that our approach outperforms SinNeRF in synthesizing novel-view images with better structures and textures.

\begin{table}[]
\caption{Quantitative comparison of our method against other state-of-the-art methods  on the NeRF Synthetic dataset.}
\centering
\scalebox{0.9}{
\begin{tabular}{lllllll}
\hline
          & \multicolumn{3}{l}{Lego}                            & \multicolumn{3}{l}{Hotdog}                          \\ \cline{2-7} 
Method    & SSIM$\uparrow$ & PSNR$\uparrow$ & LPIPS$\downarrow$ & SSIM$\uparrow$ & PSNR$\uparrow$ & LPIPS$\downarrow$ \\ \hline
DS-NeRF   & 0.77           & 16.62          & 0.17              & 0.67           & 14.16          & 0.29              \\
DietNeRF  & 0.72           & 15.07          & 0.21              & 0.69           & 16.28          & 0.26              \\
PixelNeRF & 0.72           & 14.25          & 0.22              & 0.71           & 16.67          & 0.24              \\
SinNeRF   & 0.82           & 20.97          & \textbf{0.09}     & 0.77           & 19.78          & 0.17              \\
AdaMPI    & 0.82           & 19.20          & 0.20              & 0.82           & 18.40          & 0.19              \\
Ours      & \textbf{0.87}  & \textbf{22.99} & \textbf{0.09}     & \textbf{0.84}  & \textbf{22.29} & \textbf{0.12}     \\ \hline
\end{tabular}
}
\label{tab: synthetic}
\end{table}

\subsection{Ablation Study}

\subsubsection{Inpainting Module}
\begin{figure*}
  \includegraphics[width=0.9\textwidth]{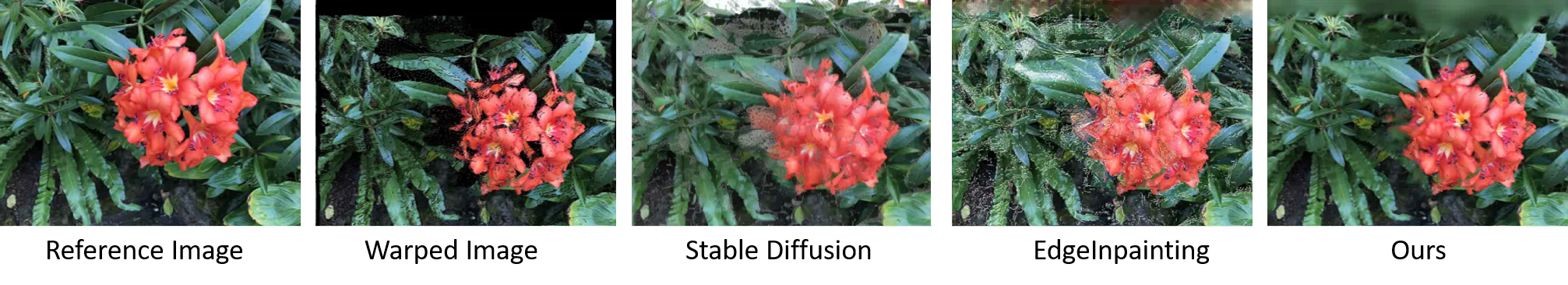}
  \caption{Comparison of different inpainting approaches. Our depth-aware inpainter generates more realistic results.}
  \label{fig:fig_ablation_inpaint}
\end{figure*}

While Stable Diffusion~\cite{rombach2021highresolution} is effective for outpainting in our approach, it is not 3D-aware, which can lead to unstable and flickering contents when the camera moves. Additionally, Stable Diffusion struggles to inpaint highly irregular missing areas due to its pre-training routine. Furthermore, finetuning the pretrained Stable Diffusion model on a single image is not feasible in our case.
Fig.~\ref{fig:fig_ablation_inpaint} presents a comparison of different inpainting methods. Our approach achieves the best results. In contrast, Stable Diffusion~\cite{rombach2021highresolution} struggles to inpaint complex missing areas in this task.

\subsubsection{MPI Module}

Table~\ref{tab: ablation MPI} demonstrates that our method achieves higher image quality in the synthesis results based on MPI compared to our pseudo-novel-view data generated by the depth-aware inpainter. Our qualitative results further support the effectiveness of MPI in stabilizing inpainted areas that lack explicit 3D geometry, resulting in more realistic outcomes with improved 3D consistency.

\begin{figure*}
  \includegraphics[width=0.9\textwidth]{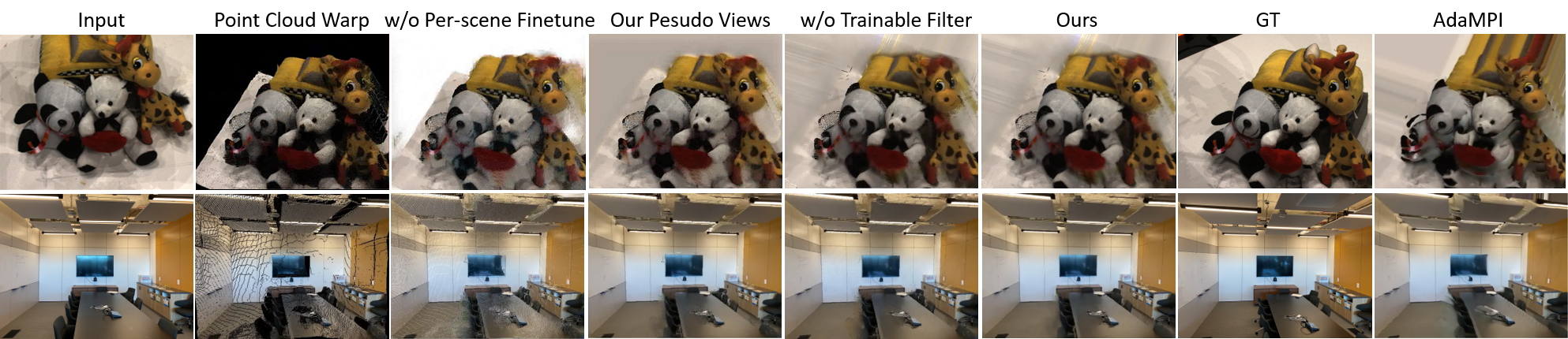}
  \caption{The ablation study on DTU scan 84.}
  \label{fig:fig_ablation_scan84}
\end{figure*}

In Fig.~\ref{fig:fig_ablation_scan84}, we show the qualitative ablation study conducted to investigate our method. Starting with the reference image and its estimated depth from DTU scan 84, we employ point cloud perspective warp to render the novel view together with depth information. Black holes/areas in the rendered image indicate missing geometry and our depth-aware inpainter completes both the image and the depth to generate pseudo views. Guided by these inpainted pseudo views, we fit the expanded MPI to obtain the final results. Our finetune strategy improves the quality of inpainted textures by resolving issues such as misalignment with the pretrained routine and ignoring small missing areas in the input data. Our trainable filter effectively removes noisy points in the pseudo views caused by points projection and generates smoother views and accelerates the optimization process by sharing information between rays.

Table~\ref{tab: ablation MPI} presents the ablation study, where we evaluate the performance of our method without different major components to assess the impact of these design choices on the final metrics. 

\begin{table}[]
\caption{The ablation study on DTU scan 84. PF, PV, and TF refer to per-scene finetuning, pseudo-views, and trainable filters, respectively.}
\centering
\scalebox{0.9}{
\begin{tabular}{lllllll}
\hline
                 & \multicolumn{3}{c}{DTU Scan-84}                     & \multicolumn{3}{c}{LLFF Room}                       \\ \cline{2-7} 
Method           & SSIM$\uparrow$ & PSNR$\uparrow$ & LPIPS$\downarrow$ & SSIM$\uparrow$ & PSNR$\uparrow$ & LPIPS$\downarrow$ \\ \hline
AdaMPI           & 0.70           & \textbf{19.03} & 0.39              & 0.68           & 16.04          & \textbf{0.34}     \\ \hline
w/o PF     & 0.67           & 17.96          & 0.41              & 0.66           & 16.40          & 0.37              \\
Our PV & 0.69           & 18.11          & 0.39              & 0.67           & 16.86          & \textbf{0.34}     \\ \hline
w/o TF       & 0.75           & 18.23          & 0.39              & 0.68           & 16.60          & 0.35              \\
Ours             & \textbf{0.76}  & 18.72          & \textbf{0.38}     & \textbf{0.70}  & \textbf{17.20} & \textbf{0.34}     \\ \hline
\end{tabular}
}
\label{tab: ablation MPI}
\end{table}

\section{Limitations}
As a common issue for single-image novel view synthesis methods, our SinMPI heavily relies on the accuracy of the depth estimator and cannot fix depth estimation errors. Additionally, similar to other MPI-based approaches, SinMPI performs poorly on slanted surfaces. Furthermore, the inpainter used in SinMPI struggles to generate realistic textures when large missing areas exist.

View-dependent effects are important for realistic rendering, but our method cannot model view-dependent effects such as specular reflections, due to the challenge of inferring light conditions from a single image.
We leave this as our future work.

\section{Conclusion}
In this paper, we proposed SinMPI, a novel single-image approach for synthesizing novel views that constructs an expanded multiplane space from a single input image, enabling fast and 3D-consistent view synthesis. Our method significantly expands the perspective range of MPIs and thus allows a wider range of camera movement. Extensive qualitative and quantitative evaluations across various datasets demonstrated that our method outperforms other state-of-the-art methods in single-image novel view synthesis. \looseness=-1

\begin{acks}
This work was supported by National Natural Science
Foundation of China (Grant No.: 62372015), Center For Chinese Font Design and Research, and Key Laboratory of Intelligent Press Media Technology.
\end{acks}


\bibliographystyle{ACM-Reference-Format}
\bibliography{references}

\clearpage
\pagebreak

\pagebreak

\end{document}